\documentclass[11pt,a4paper]{article}
\usepackage[dvipsnames]{xcolor}
\usepackage{graphicx}
\usepackage{amsmath}
\usepackage{multirow}
\usepackage{array}
\usepackage{caption}

\usepackage[hyperref]{naaclhlt2019}
\usepackage{times}
\usepackage{latexsym}

\usepackage{url}

\aclfinalcopy
\def\aclpaperid{458} 

\newcolumntype{L}[1]{>{\raggedright\let\newline\\\arraybackslash\hspace{0pt}}m{#1}}
\newcolumntype{C}[1]{>{\centering\let\newline\\\arraybackslash\hspace{0pt}}m{#1}}
\newcolumntype{R}[1]{>{\raggedleft\let\newline\\\arraybackslash\hspace{0pt}}m{#1}}
\def\doubleunderline#1{\underline{\underline{#1}}}

\title{Question Answering as an Automatic Evaluation Metric for News Article Summarization}
  
\author{
Matan Eyal\textsuperscript{\normalfont 1, 2}, 
Tal Baumel\textsuperscript{\normalfont 1, 3}, 
Michael Elhadad\textsuperscript{\normalfont 1} \\
\textsuperscript{1}Dept. Computer Science, Ben Gurion University \\
\textsuperscript{2}IBM Research, Israel, \textsuperscript{3}Microsoft\\
  {\tt \{mataney, elhadad\}@cs.bgu.ac.il,
  tabaumel@microsoft.com}}

\date{}

\begin{document}
\maketitle

\begin{abstract}
  Recent work in the field of automatic summarization and headline generation focuses on maximizing ROUGE scores for various news datasets. We present an alternative, extrinsic, evaluation metric for this task, \textit{\textbf{A}nswering \textbf{P}erformance for \textbf{E}valuation of \textbf{S}ummaries}. APES utilizes recent progress in the field of reading-comprehension to quantify the ability of a summary to answer a set of manually created questions regarding central entities in the source article. We first analyze the strength of this metric by comparing it to known manual evaluation metrics. We then present an end-to-end neural abstractive model that maximizes APES, while increasing ROUGE scores to competitive results.
\end{abstract}
\section{Introduction}

\begin{figure}[t!]
\fontsize{8pt}{10pt}\selectfont
\begin{center}
\begin{tabular}{|p{\dimexpr\linewidth-2.5\tabcolsep}|}
\hline 
\textbf{\citet{see2017get}'s Summary:}
bolton will offer new contracts to emile heskey, 37, eidur gudjohnsen, 36, and adam bogdan, 27. heskey and gudjohnsen joined on short-term deals in december. eidur gudjohnsen has scored five times in the championship .
\\
\textbf{APES score: 0.33}
\\
\hline
\textbf{Baseline Model Summary (Encoder / Decoder / Attention / Copy / Coverage):} bolton will offer new contracts to emile heskey, 37, eidur gudjohnsen, 36, and goalkeeper adam bogdan, 27. heskey and gudjohnsen joined on short-term deals in december, and have helped neil lennon 's side steer clear of relegation. eidur gudjohnsen has scored five times in the championship, as well as once in the cup this season .
\\
\textbf{APES score: 0.33}
\\
\hline
\textbf{Our Model (APES optimization):} bolton will offer new contracts to emile heskey, 37, eidur gudjohnsen, 36, and goalkeeper adam bogdan, 27. heskey joined on short-term deals in december, and have helped neil lennon 's side steer clear of relegation. eidur gudjohnsen has scored five times in the championship, as well as once in the cup this season. lennon has also fined midfielders barry bannan and neil danns two weeks wages this week. both players have apologised to lennon .
\\
\textbf{APES score: 1.00}
\\
\hline

\textbf{Questions from the CNN/Daily Mail Dataset:}\\

\textbf{Q}: goalkeeper \underline{\hspace{0.6cm}} also rewarded with new contract; \textbf{A}: \textcolor{Green}{adam bogdan}
\\
\textbf{Q}: \underline{\hspace{0.6cm}} and neil danns both fined by club after drinking incident; \textbf{A}: \textcolor{Green}{barry bannan}
\\
\textbf{Q}: barry bannan and \underline{\hspace{0.6cm}} both fined by club after drinking incident; \textbf{A}: \textcolor{Green}{neil danns}
\\
\hline
\end{tabular}
\end{center}
\caption{Example 3083 from the test set.}
\label{figure:ex1}
\end{figure}

The task of automatic text summarization aims to produce a concise version of a source document while preserving its central information. Current summarization models are divided into two approaches, \textit{extractive} and \textit{abstractive}. In extractive summarization, summaries are created by selecting a collection of key sentences from the source document (\textit{e.g.}, \citet{nallapati2017summarunner}; \citet{narayan2018ranking}). Abstractive summarization, on the other hand, aims to rephrase and compress the input text in order to create the summary. Progress in sequence-to-sequence models \cite{sutskever2014sequence} has led to recent success in abstractive summarization models. Current models \cite{nallapati2016abstractive,see2017get,paulus2017deep,celikyilmaz2018deep} made various adjustments to sequence-to-sequence models to gain improvements in ROUGE \cite{rouge} scores. 

ROUGE has achieved its status as the most common method for summaries evaluation by showing high correlation to manual evaluation methods, \textit{e.g.}, the Pyramid method \cite{pyramid}. Tasks like TAC AESOP \cite{tac11} used ROUGE as a strong baseline and confirmed the correlation of ROUGE with manual evaluation.

While it has been shown that ROUGE is correlated to Pyramid, \citet{louis2013automatically} show that this summary level correlation decreases significantly when only a single reference is given. In contrast to the smaller manually curated DUC datasets used in the past, more recent large-scale summarization and headline generation datasets (\textit{CNN/Daily Mail} \cite{hermann2015teaching}, Gigaword \citep{graff2003english}, New York Times \cite{sandhaus2008new}) provide only a single reference summary for each source document. In this work, we introduce a new automatic evaluation metric more suitable for such single reference news article datasets.

We define APES, \textit{\textbf{A}nswering \textbf{P}erformance for \textbf{E}valuation of \textbf{S}ummaries}, a new metric for automatically evaluating summarization systems by querying summaries with a set of questions central to the input document (see Fig. \ref{figure:ex1}).

Reducing the task of summaries evaluation to an extrinsic task such as question answering is intuitively appealing. This reduction, however, is effective only under specific settings: (1) Availability of questions focusing on central information and (2) availability of a reliable question answering (QA) model.

Concerning issue 1, questions focusing on salient entities can be available as part of the dataset: the headline generation dataset most used in recent years, the \textit{CNN/Daily Mail} dataset \cite{hermann2015teaching}, was constructed by creating questions about entities that appear in the reference summary. Since the target summary contains salient information from the source document, we consider all entities appearing in the target summary as \textit{salient entities}.
In other cases, salient questions can be generated in an automated manner, as we discuss below.  


Concerning issue 2, we focus on a relatively easy type of questions: given source documents and associated questions, a QA system can be trained over fill-in-the-blank type questions as was shown in \citet{hermann2015teaching} and \citet{chen2016thorough}. In their work, \citet{chen2016thorough} achieve `ceiling performance' for the QA task on the \textit{CNN/Daily Mail} dataset. We empirically assess in our work whether this performance level (accuracy of 72.4 and 75.8 over \textit{CNN} and \textit{Daily Mail} respectively) makes our evaluation scheme feasible and well correlated with manual summary evaluation.

Given the availability of salient questions and automatic QA systems, we propose APES as an evaluation metric for news article datasets, the most popular summarization genre in recent years. 

\begin{figure}
    \includegraphics[scale=0.45]{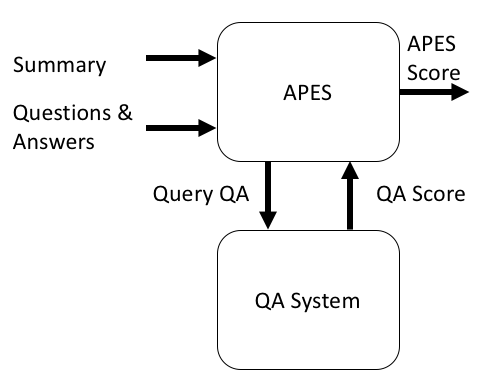}
    \caption{Evaluation flow of APES.}
\label{figure:reduction}
\end{figure}

To measure the APES metric of a candidate summary, we run a trained QA system with the summary as input alongside a set of questions associated with the source document. The APES metric for a summarization model is the percentage of questions that were answered correctly over the whole dataset, as depicted in Fig.~\ref{figure:reduction}. We leave the task of extending this method to other genres for future work.

Our contributions in this work are: (1) We first present APES, a new extrinsic summarization evaluation metric; (2) We show APES strength through an analysis of its correlation with Pyramid and Responsiveness manual metrics; (3) we present a new abstractive model which maximizes APES by increasing attention scores of salient entities, while increasing ROUGE to competitive level. We make two software packages available online: (a) An evaluation library which receives the same input as ROUGE and produces both APES and ROUGE scores.\footnote{\url{www.github.com/mataney/APES}} (b) Our PyTorch \cite{paszke2017automatic} based summarizer that optimizes APES scores together with trained models.\footnote{\url{www.github.com/mataney/APES-optimizer}}


\section{Related Work}

\subsection{Evaluation Methods}

Automatic evaluation metrics of summarization methods can be categorized into either \textit{intrinsic} or \textit{extrinsic} metrics. Intrinsic metrics measure a summary's quality by measuring its similarity to a manually produced target gold summary or by inspecting properties of the summary. Examples of such metrics include ROUGE \cite{rouge}, Basic Elements \cite{hovy2006automated} and Pyramid \cite{pyramid}. Alternatively, extrinsic metrics test the ability of a summary to support performing related tasks and compare the performance of humans or systems when completing a task that requires understanding the source document \cite{steinberger2012evaluation}. Such extrinsic tasks may include text categorization, information retrieval, question answering \cite{jing1998summarization} or assessing the relevance of a document to a query \cite{hobson2007task}.

ROUGE, or ``Recall-Oriented Understudy for Gisting Evaluation'' \cite{rouge}, refers to a set of automatic intrinsic metrics for evaluating automatic summaries. ROUGE-N scores a candidate summary by counting the number of N-gram overlaps between the automatic summary and the reference summaries. Other notable metrics from this family are ROUGE-L, where scores are given by the Longest Common Subsequence (LCS) between the suggested and reference documents, and ROUGE-SU4, which uses skip-bigram, a more flexible method for computing the overlap of bigrams.

The Pyramid method \cite{pyramid} is a manual evaluation metric that analyzes multiple human-made summaries into ``Summary Content Units'' (SCUs) and assigns importance weights to each SCU. Different summaries are scored by assessing the extent to which they convey SCUs according to their respective weights. Pyramid is most effective when multiple human-made summaries alongside manual intervention to detect SCUs in source and target documents. The Basic Elements method \cite{hovy2006automated}, an automated procedure for finding short fragments of content, has been suggested to automate a method related to Pyramid. Like Pyramid, this method requires multiple human-made gold summaries, making this method expensive in time and cost. Responsiveness \cite{respon}, another manual metric is a measure of overall quality combining both content selection, like Pyramid, and linguistic quality. Both Pyramid and Responsiveness are the standard manual approaches for content evaluation of summaries.

Automated Pyramid evaluation has been attempted in the past \cite{Owczarzak2009DEPEVALsummDE, Yang2016PEAKPE,Hirao2018AutomaticPE}.  This task is complex because it requires (1) identifying SCUs in a text, which requires syntactic parsing and the extraction of key subtrees from the identified units, and (2) the clustering of these extracted textual elements into semantically similar SCUs.  These two operations are noisy, and the compounded performance summary evaluation is relying on noisy intermediary representation accordingly suffers.

Other relevant quantities for summaries quality assessment include: readability (or fluency), grammaticality, coherence and structure, focus, referential clarity, and non-redundancy. Although some automatic methods were suggested as summarization evaluation metrics \cite{vadlapudi2010automated, tay2017skipflow}, these metrics are commonly assessed manually, and, therefore, rarely reported as part of experiments.

Our proposed evaluation method, APES, attempts to capture the capability of a summary to enable readers to answer questions -- similar to the manual task initially discussed in \citet{jing1998summarization} and recently reported in \citet{narayan2018ranking}. Our contribution consists of automating this method and assessing the feasibility of the resulting approximation.

\subsection{Neural Methods for Abstractive and Extractive Summarization} 

The first paper to use an end-to-end neural network for the summarization task was \citet{rush2015neural}: this work is based on a sequence-to-sequence model \cite{sutskever2014sequence} augmented with an attention mechanism \cite{bahdanau2014neural}. \citet{nallapati2016abstractive} was the first to tackle the headline generation problem using the \textit{CNN/Daily Mail} dataset \cite{hermann2015teaching} adopted for the summarization task. 

\citet{see2017get} followed the work of \citet{nallapati2016abstractive} and added an additional loss term to reduce repetitions at decoding time. \citet{paulus2017deep} introduces intra-attention in order to attend over both the input and previously generated outputs. The authors also present a hybrid learning objective designed to maximize ROUGE scores using Reinforcement Learning.

All the papers mentioned above have been evaluated using ROUGE, and all, except for \citet{rush2015neural}, used \textit{CNN/Daily Mail} as their main headline generation dataset. Of all the mentioned models we compare our suggested model only to \citep{see2017get}, as it is the only paper to publish output summaries.
\section{APES}
\label{sec:apes}

Evaluating a summarization system with APES applies the following method: APES receives a set of news articles summaries, question-and-answer pairs referring to central information from the text and an automatic QA system. Then, APES uses this QA system to determine the total number of questions answered correctly according to the received summaries. The evaluation process is depicted in Fig.~\ref{figure:reduction}. We use \citet{chen2016thorough}'s model trained on the \textit{CNN} dataset as our QA system for all our experiments.  For a given summarizer and a given dataset, APES reports the average number of questions correctly answered from the summaries produced by the system.

This method is especially relevant for the main headline generation dataset used in recent years, the \textit{CNN/Daily Mail} dataset, as it was initially created for the question answering task by \citet{hermann2015teaching}.
It contains 312,085 articles with relevant questions scraped from the two news agencies' websites. The questions were created by removing different entities from the manually produced highlights to create 1,384,887 fill-in-the-blank questions. The dataset was later repurposed by \citet{cheng2016neural} and \citet{nallapati2016abstractive} to the summarization task by reconstructing the original highlights from the questions. Fig.~\ref{figure:create_questions} shows an example for creating questions out of a given summary.

\begin{figure}[t]
\fontsize{8pt}{12pt}\selectfont
\begin{center}
\begin{tabular}{|p{\dimexpr\linewidth-2\tabcolsep}|}
\hline 
\textbf{Original Reference Summary:}\\
Arsenal beat Burnley 1-0 in the EPL. a goal from Aaron Ramsey secured all three points. win cuts Chelsea 's EPL lead to four points .\\
 \hline
\textbf{Produces questions:}

\textbf{Q}: \underline{\hspace{0.6cm}} beat @entity7 1-0 in the @entity4; \textbf{A}: Arsenal

\textbf{Q}: @entity0 beat \underline{\hspace{0.6cm}} 1-0 in the @entity4; \textbf{A}: Burnley

\textbf{Q}: @entity0 beat @entity7 1-0 in the \underline{\hspace{0.6cm}}; \textbf{A}: EPL

\textbf{Q}: a goal from \underline{\hspace{0.6cm}} secured all three points; \textbf{A}: Aaron Ramsey

\textbf{Q}: win cuts \underline{\hspace{0.6cm}}'s @entity4 lead to four points; \textbf{A}: Chelsea

\textbf{Q}: win cuts @entity19 's \underline{\hspace{0.6cm}} lead to four points; \textbf{A}: EPL
\\
\hline
\end{tabular}
\end{center}
\caption{Example 202 from the \textit{CNN/Daily Mail} test set.}
\label{figure:create_questions}
\end{figure}

\subsection{Using APES as an Evaluation Metric for any News Datasets}

\label{subsec:apes_on_other_ds}

When questions are not intrinsically available, one requires to (1) automatically generate relevant questions; (2) use an appropriate automatic QA system.

Similarly to the method used in \citet{hermann2015teaching}, we produce fill-in-the-blank questions in the following way: given a reference summary, we find all possible entities, (\textit{i.e.}, Name, Nationality, Organization, Geopolitical Entity or Facility) using an NER system \cite{spacy} and we create fill-in-the-blank type questions where the answers are these entities. We provide code for this procedure and apply it on the  \textit{AESOP} datasets in our experiments\footnote{\url{https://github.com/mataney/APES-on-TAC2011}}.

For the automatic QA system, we reused in our experiment the same QA system trained on \textit{CNN/Daily Mail} for different News datasets (including AESOP).  To enable reproducibility, the trained models used are available online.
\section{APES on the TAC2011 AESOP Task}

\begin{table*}[t]
\centering
\begin{tabular}{c|c|c|c|c|c}
  & ROUGE-1 & ROUGE-2 & ROUGE-L & ROUGE-SU & APES \\
\hline
Pyramid & 0.590 & 0.468* & 0.599 & 0.563* & \textbf{0.608} \\
Responsiveness & 0.540 & 0.518* & 0.537 & 0.541 & \textbf{0.576} \\
\end{tabular}
\caption[Pearson Correlation of ROUGE and APES against Pyramid and Responsiveness on summary level]{Pearson Correlation of ROUGE and APES against Pyramid and Responsiveness on summary level. Statistically significant differences are marked with *.}
\label{table:apes_on_tac}
\end{table*}

To evaluate if an automatic metric can accurately measure a summarization system performance, we measure its correlation to manual metrics. The TAC 2011 \textit{Automatically Evaluating Summaries of Peers} (\textit{AESOP}) task \cite{tac11} has provided a dataset that includes, alongside the source documents and reference summaries, three manual metrics: Pyramid \cite{pyramid},  Overall Responsiveness \cite{respon} and Overall Readability. Two sets of documents are provided, we use only the documents from the first set (Generic summarization), as the second set is relevant to the update summarization task.

To evaluate APES on the AESOP dataset, we create the required set of questions as presented in Fig.~\ref{figure:create_questions}. We used the same QA system \cite{chen2016thorough} trained on the \textit{CNN} dataset. This system is a competent QA system for this dataset, as both AESOP and \textit{CNN} consist of news articles. Training a QA model on the AESOP dataset would be optimal, but it is not possible due to the small size of this dataset. Nonetheless, even this incomplete QA system reports valuable results that justify APES value.

While the two datasets are similar, they differ dramatically in the type of topics the articles cover. \textit{CNN/Daily Mail} articles deal with people, or more generally, Named Entities, averaging 6 named entities per summary. In contrast, TAC summaries average 0.87 entities per summary. The TAC dataset is divided into various topics. The first four topics, \textit{Accidents and Natural Disasters}, \textit{Attacks}, \textit{Health and Safety} and \textit{Endangered Resources} average 0.65 named entities per summary, making them incomparable to the typical case in the \textit{CNN/Daily Mail} dataset. The last topic, \textit{Investigations and Trials}, averages 3.35 named entities per summary, making it more similar. We report correlation only on this segment of TAC, which contains 204 documents.

We follow the work of \citet{louis2013automatically} and compare input level APES scores with manual Pyramid and Responsiveness scores provided in the AESOP task. Results are in Table~\ref{table:apes_on_tac}. In \textit{Input level}, correlation is computed for each summary against its manual score.  In contrast, \textit{system level} reports the average score for a summarization system over the entire dataset.

While ROUGE baselines were beaten only by a very small number of suggested metrics in the original AESOP task, we find that APES shows better correlation than the popular R-1, R-2 and R-L, and the strong R-SU. Although showing statistical significance for our hypothesis is difficult because of the small dataset size, we claim APES gives an additional value comparing to ROUGE: ROUGE metrics are highly correlated with each other (around 0.9) as shown in Table~\ref{table:corr_matrix}, indicating that multiple ROUGE metrics provide little additional information. In contrast, APES is not correlated with ROUGE metrics to the same extent (around 0.6). The above suggests that APES offers additional information regarding the text in a manner that ROUGE does not. For this reason, we believe APES complements ROUGE.

\begin{table}
\centering
\begin{tabular}{c|ccccc}
         & R-1  & R-2  & R-L  & R-SU & APES \\
\hline
R-1  & 1.00 & 0.83 & 0.92 & 0.94 & 0.66 \\
R-2  &      & 1.00 & 0.82 & 0.90 & 0.61 \\
R-L  &      &      & 1.00 & 0.89 & 0.66 \\
R-SU &      &      &      & 1.00 & 0.67 \\
APES &      &      &      &      & 1.00
\end{tabular}
\caption{Correlation matrix of ROUGE and APES.}
\label{table:corr_matrix}
\end{table}

\citet{louis2013automatically} further shows that ROUGE correlation to manual scores tends to drop when reducing the number of reference summaries. While APES is not immune to this, as the number of questions becomes smaller when the number of reference summaries is reduced, it still performs well when reducing the number of references to a single document. In the AESOP dataset, when comparing with respect to each of the 8 assessors separately on Pyramid and Responsiveness, the correlation of APES is highest in 7 out of 16 trials, while that of R1 is highest in 6 trials and RL in 2 trials. In general, the correlation between any of the metrics and single references is extremely noisy, indicating that reliance on evaluations of a single reference, which is standard on large-scale summarization datasets, is far from satisfactory.

We have established that APES achieves equal or improved correlation with manual metrics when compared to ROUGE, and captures a different type of information than ROUGE, by that, APES can complement ROUGE as an automatic evaluation metric. We now turn to develop a model that directly attempts to optimize APES.

\section{Model}

News articles include a high number of named entities. When analyzing systems performance on APES (Table~\ref{table:avg_entities}), a system may fail either when it misses to generate a salient entity in the summary, or when it includes the salient entity, but in a context not relevant to corresponding questions. When this happens, the QA system would not be able to identify the entity as an answer to a question referring to the context.

\begin{table}[t]
\centering
\fontsize{9pt}{12pt}\selectfont
\begin{tabular}{|C{2.2cm}|C{0.8cm} |C{1.2cm}|C{1.1cm}| }
\hline
Model & APES & \#Entities & \#Salient Entities \\
\hline
\citet{see2017get} & 38.2 & 4.90 & 2.57 \\
Baseline model & 39.8 & 4.99 & 2.61 \\
\hline
Gold Summaries & 85.5 & 6.00 & 4.90\\
\hline
\end{tabular}
\caption{Average number of entities and salient entities.}
\label{table:avg_entities}
\end{table}

We compared the average number and type of entities in summaries generated by existing automatic summarizers to that in reference summaries. We note that the observed models, while producing state-of-the-art ROUGE scores and a high number of named entities (5 vs. 6 on average), fail to focus on salient entities when generating a summary (about 2.6 salient entities are mentioned on average vs. 4.9 in the reference summaries). Notice that solely increasing the number of entities is damaging: mentioning too many entities causes a decrease in the QA accuracy, as the number of possible answers increases, which would distract the QA system. This has motivated us in suggesting the following model.

\subsection{Baseline Model}

To experiment with direct optimization of APES, we reconstruct as a starting point a model that encapsulates the key techniques used in recent abstractive summarization models. Our model is based on the OpenNMT project \cite{opennmt}. All PyTorch \cite{paszke2017automatic} code, including entities attention and beam search refinement is available online\footnote{\url{www.github.com/mataney/APES-optimizer}}. We also include generated summaries and trained models in this repository.

Recent work in the field of abstractive summarization \cite{rush2015neural,nallapati2016abstractive,see2017get,paulus2017deep} share a common architecture as the foundation for their neural models: an encoder-decoder model \cite{sutskever2014sequence} with an attention mechanism \cite{bahdanau2014neural}. \citet{nallapati2016abstractive} and \citet{see2017get} augment this model with a copy mechanism \cite{vinyals2015pointer}. This architecture minimizes the following loss function:

\begin{equation}
\label{eqn:loss}
\begin{aligned}
\textrm{loss}_t = -\log P(w^*_t) \\
\textrm{loss} = \frac{1}{T_y}\sum_{t=0}^{T_y} \textrm{loss}_t
\end{aligned}
\end{equation}

$loss_t$, is the negative log likelihood of generating the gold target word $w^*_t$ at timestep  $t$ where $P(\cdot)$ is the probability distribution over the vocabulary. We refer the reader to \citet{see2017get} for a more detailed description of this architecture.  

Unlike \citet{see2017get}, we do not train a specific coverage mechanism to avoid repetitions. Instead, we incorporate \citet{wu2016google}'s refinements of beam search in order to manipulate both the summaries' coverage and their length. In the standard beam search, we search for a sequence $Y$ that maximizes a score function $s(Y, X) = \log(P(Y|X))$. \citet{wu2016google} introduce two additional regularization factors, \textit{coverage penalty} and \textit{length penalty}. These two penalties, with an additional refinement suggested in \citet{gehrmann2018bottom}, yield the following score function:

\begin{equation}
\label{eqn:beam}
\begin{aligned}
    s(Y,X) &= \log(P(Y|X))/lp(Y) - cp(X;Y) \\
    lp(Y) &= \frac{(5+|Y|)^\alpha}{(5+1)^\alpha}\\
    cp(X;Y) &= \beta (-T_X + \sum_{i=1}^{T_X}\max(\sum_{j=1}^{T_Y}a_{i,j}, 1.0))
\end{aligned}
\end{equation}

\noindent where $\alpha, \beta$ are hyper-parameters that control the length and coverage penalties respectively and $a_{i,j}$ is the attention probability of the $j$-th target word on the $i$-th source word.

$cp(X;Y)$, the coverage penalty, is designed to discourage repeated attention to the same source word and favor summaries that cover more of the source document with respect to the attention distribution.

$lp(Y)$, the length normalization, is designed to compare between beam hypotheses of different length accurately. In general, beam search favors shorter outputs as log-probability is added at each step, yielding lower scores for longer sequences. $lp$ compensates for this tendency.

In the following section, we describe how we extend this baseline model in order to maximize the APES metric. The new model learns to incorporate more of the salient entities from the source document in order to optimize its APES metric.

\begin{table*}
\centering
\resizebox{0.8\textwidth}{!}{
\begin{tabular}{ccccc}
  \multirow{2}{*}{\textbf{Model}} & \multirow{2}{*}{\textbf{APES}}  & \multicolumn{3}{c}{\textbf{ROUGE}} \\
  &  & \textbf{1} & \textbf{2} & \textbf{L} \\
  \hline
  Source & 61.1 & - & - & -\\
Gold-Summaries & 85.5  & 100 & 100 & 100\\
Shuffled Gold-Summaries & 30.9 & 100 & 7.0 & 58.3\\
Lead 3  & 45.1 & 40.1 & 17.3 & 36.3 \\
\hline
Pointer-generator + coverage \cite{see2017get}$^*$ & 38.2 & 39.3 & 16.9 & 35.7 \\
Baseline model & 39.8 & 39.3 & 17.3 & 36.3 \\
\textbf{Our model} & \textbf{46.1} & \textbf{40.2} & \textbf{17.7} & \textbf{37.0}\\
\hline
Our model with gold entities positions  & 46.3 & 40.4 & 17.8 & 37.3
\end{tabular}
}\caption{APES: Percent of questions answered correctly using by document. *Obtained from the model uploaded to \url{github.com/abisee/pointer-generator}.}
\label{table:apes_table}
\end{table*}

\subsection{Entities Attention Layer}

As we observed, failure to capture salient entities in summaries is one cause for low APES score. To drive our model towards the identification and mention of salient entities from the source document, we introduce an additional attention layer that learns the important entities of a source document. We hypothesize that these entities are more likely to appear in the target summary, and thus are better candidate answers to one of the salient questions for this document.

We learn for each word in the source document its probability of belonging to a salient entity mention. We adopt the classical soft attention mechanism of \citet{bahdanau2014neural}: after encoding the source document, we run an additional single alignment model with an empty query and a sigmoid layer instead of the standard softmax layer. 

\begin{equation}
\begin{aligned}
a_{j}^e &= \sigma(e_{j}^e) \\
e_{j}^e &= v^T\tanh(Uh_j + b)
\end{aligned}
\end{equation}
\noindent where $U, b, v$ are learnable weight matrices, $h_j$ is the encoder hidden state for the $j$-th word and $\sigma(\cdot)$ is a logistic sigmoid function. $a_j^e$ reflects the probability of the $j$-th token of being a salient entity.

The second modification comparing to \citet{bahdanau2014neural} is that we replace the softmax function with a sigmoid: while in the standard alignment model, we intend to obtain a normalized probability distribution over all the tokens of the source document, here we would like to get a probability of each token being a salient entity independently of other tokens. In order to drive this attention layer towards salient entities, we define an additional term in the loss function.
\begin{equation}
loss_e = BCE(a^e, s^*)
\end{equation}
\noindent where $s^*$ is a binary vector of source length size, where $s_j^*=1$ if $x_j$ is a salient entity, and $0$ otherwise, and $BCE$ is the binary cross entropy function. This term is added to the standard log-likelihood loss, changing equation \eqref{eqn:loss} to the following composite loss function:
\begin{equation}\label{eqn:hybrid_loss}
\textrm{loss} = \delta\ loss_e + (1-\delta)\frac{1}{T_y}\sum_{t=0}^{T_y} \textrm{loss}_t
\end{equation}
\noindent where $\delta$ is a hyper-parameter. We join these two terms in the loss function in order to learn the entities attention layer while keeping the summarization ability learned by Eq.~\eqref{eqn:loss}.

\subsection{Entities Attention and Beam Search}

After the attention layer has learned the probability of each source token to belong to a salient entity, we pass the predicted alignment to the beam search component at test-time. Using this alignment data, we wish to encourage beam search to favor hypotheses attending salient entities.

Accordingly, we introduce a new term $ep$ to the beam search score function of equation \eqref{eqn:beam}:
\begin{equation}
\label{eqn:my_beam_pen}
\begin{aligned}
    s(Y,X) &= \log(P(Y|X))/lp(Y) - cp(X;Y) \\ &- ep(X;Y) \\
    ep(X;Y) &= \gamma \sum_{i=1}^{T_X}\max(a^e_i-\sum_{j=1}^{T_Y}a_{i,j}, 0.0)
\end{aligned}
\end{equation}

$ep(X;Y)$ penalizes summaries that do not attend parts of the source document we believe are central.

Fig.~\ref{figure:example_with_attention} compares summaries produced by this model and the baseline model by showing their respective attention distribution and the impact on the decision of which words to include in the summary based on the attention level derived from salient entities.

\section{Results}

\begin{figure*}[t]
\fontsize{7.5pt}{10pt}\selectfont
\begin{center}
\begin{tabular}{|p{\dimexpr\linewidth-2\tabcolsep}|}
\hline 
\textbf{Source document:}\\
\fontsize{7.5pt}{6pt}\selectfont
{\colorbox[cmyk]{0,0,0,0}{jack}} {\colorbox[cmyk]{0,0,0,0}{wilshere}} {\colorbox[cmyk]{0,0,0,0}{may}} {\colorbox[cmyk]{0,0,0,0}{rub}} {\colorbox[cmyk]{0,0,0,0}{shoulders}} {\colorbox[cmyk]{0,0,0,0}{with}} {\colorbox[cmyk]{0,0,0,0}{the}} {\colorbox[cmyk]{0,0,0,0}{likes}} {\colorbox[cmyk]{0,0,0,0}{of}} {\colorbox[cmyk]{0,0,0,0}{alexis}} {\colorbox[cmyk]{0,0,0,0}{sanchez}} {\colorbox[cmyk]{0,0,0,0}{and}} {\colorbox[cmyk]{0,0,0,0}{mesut}} {\colorbox[cmyk]{0,0,0,0}{ozil}} {\colorbox[cmyk]{0,0,0,0}{on}} {\colorbox[cmyk]{0,0,0,0}{a}} {\colorbox[cmyk]{0,0,0,0}{daily}} {\colorbox[cmyk]{0,0,0,0}{basis}} {\colorbox[cmyk]{0,0,0,0}{but}} {\colorbox[cmyk]{0,0,0,0}{he}} {\colorbox[cmyk]{0,0,0,0}{was}} {\colorbox[cmyk]{0,0,0,0}{left}} {\colorbox[cmyk]{0,0,0,0}{starstruck}} {\colorbox[cmyk]{0,0,0,0}{on}} {\colorbox[cmyk]{0,0,0,0}{thursday}} {\colorbox[cmyk]{0,0,0,0}{evening}} {\colorbox[cmyk]{0,0,0,0}{when}} {\colorbox[cmyk]{0,0,0,0}{he}} {\colorbox[cmyk]{0,0,0,0}{met}} {\colorbox[cmyk]{0,0,0,0}{brazil}} {\colorbox[cmyk]{0,0,0,0}{legend}} {\colorbox[cmyk]{0,0,0,0}{pele}} {\colorbox[cmyk]{0,0,0,0}{.}} {\colorbox[cmyk]{0,0,0,0}{even}} {\colorbox[cmyk]{0,0,0,0}{better}} {\colorbox[cmyk]{0,0,0,0}{for}} {\colorbox[cmyk]{0,0,0,0}{wilshere}} {\colorbox[cmyk]{0,0,0,0}{,}} {\colorbox[cmyk]{0,0,0,0}{the}} {\colorbox[cmyk]{0,0,0,0}{arsenal}} {\colorbox[cmyk]{0,0,0,0}{midfielder}} {\colorbox[cmyk]{0,0,0,0}{was}} {\colorbox[cmyk]{0,0,0,0}{given}} {\colorbox[cmyk]{0,0,0,0}{the}} {\colorbox[cmyk]{0,0,0,0}{opportunity}} {\colorbox[cmyk]{0,1.0,1.0,0}{\underline{to}}} {\colorbox[cmyk]{0,1.0,1.0,0}{\underline{interview}}} {\colorbox[cmyk]{0,0,0,0}{the}} {\colorbox[cmyk]{0,1.0,1.0,0.35}{\underline{three-time}}} {\colorbox[cmyk]{0,1.0,1.0,0}{\underline{world}}} {\colorbox[cmyk]{0,1.0,1.0,0}{\underline{cup}}} {\colorbox[cmyk]{0,1.0,1.0,0}{\underline{winner}}} {\colorbox[cmyk]{0,1.0,1.0,0.35}{\underline{during}}} {\colorbox[cmyk]{0,0,0,0}{the}} {\colorbox[cmyk]{0,0,0,0}{launch}} {\colorbox[cmyk]{0,0,0,0}{party}} {\colorbox[cmyk]{0,0,0,0}{of}} {\colorbox[cmyk]{0,0,0,0}{10ten}} {\colorbox[cmyk]{0,0,0,0}{talent}} {\colorbox[cmyk]{0,0,0,0}{.}} {\colorbox[cmyk]{0,0,0,0}{both}} {\colorbox[cmyk]{0,1.0,1.0,0.09}{\underline{wilshere}}} {\colorbox[cmyk]{0,1.0,1.0,0}{\underline{and}}} {\colorbox[cmyk]{0,1.0,1.0,0}{\underline{pele}}} {\colorbox[cmyk]{0,0,0,0}{,}} {\colorbox[cmyk]{0,0,0,0}{along}} {\colorbox[cmyk]{0,0,0,0}{with}} {\colorbox[cmyk]{0,0,0,0}{glenn}} {\colorbox[cmyk]{0,0,0,0}{hoddle}} {\colorbox[cmyk]{0,0,0,0}{,}} {\colorbox[cmyk]{0,1.0,1.0,0.25}{\underline{are}}} {\colorbox[cmyk]{0,1.0,1.0,0}{\underline{clients}}} {\colorbox[cmyk]{0,1.0,1.0,0}{\underline{and}}} {\colorbox[cmyk]{0,1.0,1.0,0.29}{\underline{the}}} {\colorbox[cmyk]{0,1.0,1.0,0.31}{\underline{england}}} {\colorbox[cmyk]{0,1.0,1.0,0}{\underline{international}}} {\colorbox[cmyk]{0,1.0,1.0,0.15}{\underline{made}}} {\colorbox[cmyk]{0,0,0,0}{sure}} {\colorbox[cmyk]{0,0,0,0}{his}} {\colorbox[cmyk]{0,0,0,0}{fans}} {\colorbox[cmyk]{0,0,0,0}{on}} {\colorbox[cmyk]{0,0,0,0}{twitter}} {\colorbox[cmyk]{0,0,0,0}{knew}} {\colorbox[cmyk]{0,0,0,0}{about}} {\colorbox[cmyk]{0,0,0,0}{their}} {\colorbox[cmyk]{0,0,0,0}{meeting}} {\colorbox[cmyk]{0,0,0,0}{by}} {\colorbox[cmyk]{0,0,0,0}{posting}} {\colorbox[cmyk]{0,0,0,0}{several}} {\colorbox[cmyk]{0,0,0,0}{tweets}} {\colorbox[cmyk]{0,0,0,0}{.}} {\colorbox[cmyk]{0,0,0,0}{brazil}} {\colorbox[cmyk]{0,0,0,0}{legend}} {\colorbox[cmyk]{0,0,0,0}{pele}} {\colorbox[cmyk]{0,0,0,0}{-lrb-}} {\colorbox[cmyk]{0,0,0,0}{left}} {\colorbox[cmyk]{0,0,0,0}{-rrb-}} {\colorbox[cmyk]{0,0,0,0}{and}} {\colorbox[cmyk]{0,0,0,0}{arsenal}} {\colorbox[cmyk]{0,0,0,0}{midfielder}} {\colorbox[cmyk]{0,0,0,0}{jack}} {\colorbox[cmyk]{0,0,0,0}{wilshere}} {\colorbox[cmyk]{0,0,0,0}{pose}} {\colorbox[cmyk]{0,0,0,0}{for}} {\colorbox[cmyk]{0,0,0,0}{a}} {\colorbox[cmyk]{0,0,0,0}{photo}} {\colorbox[cmyk]{0,0,0,0}{during}} {\colorbox[cmyk]{0,0,0,0}{launch}} {\colorbox[cmyk]{0,0,0,0}{of}} {\colorbox[cmyk]{0,0,0,0}{10ten}} {\colorbox[cmyk]{0,0,0,0}{talent}} {\colorbox[cmyk]{0,0,0,0}{.}} {\colorbox[cmyk]{0,0,0,0}{wilshere}} {\colorbox[cmyk]{0,0,0,0}{was}} {\colorbox[cmyk]{0,0,0,0}{given}} {\colorbox[cmyk]{0,0,0,0}{the}} {\colorbox[cmyk]{0,0,0,0}{`}} {\colorbox[cmyk]{1.0,0,1.0,0.01}{\doubleunderline{honour}}} {\colorbox[cmyk]{1.0,0,1.0,0.03}{\doubleunderline{to}}} {\colorbox[cmyk]{1.0,0,1.0,0.14}{\doubleunderline{interview}}} {\colorbox[cmyk]{1.0,0,1.0,0.47}{\doubleunderline{the}}} {\colorbox[cmyk]{1.0,0,1.0,0.47}{\doubleunderline{legendary}}} {\colorbox[cmyk]{1.0,0,1.0,0.02}{\doubleunderline{pele}}} {\colorbox[cmyk]{1.0,0,1.0,0.4}{\doubleunderline{and}}} {\colorbox[cmyk]{1.0,0,1.0,0.06}{\doubleunderline{asked}}} {\colorbox[cmyk]{1.0,0,1.0,0.53}{\doubleunderline{twitter}}} {\colorbox[cmyk]{1.0,0,1.0,0}{\doubleunderline{questions}}} {\colorbox[cmyk]{1.0,0,1.0,0}{\doubleunderline{from}}} {\colorbox[cmyk]{1.0,0,1.0,0.13}{\doubleunderline{fans}}} {\colorbox[cmyk]{1.0,0,1.0,0.35}{\doubleunderline{.}}} {\colorbox[cmyk]{0,0,0,0}{earlier}} {\colorbox[cmyk]{0,0,0,0}{on}} {\colorbox[cmyk]{0,0,0,0}{thursday}} {\colorbox[cmyk]{0,0,0,0}{,}} {\colorbox[cmyk]{0,0,0,0}{wilshere}} {\colorbox[cmyk]{0,0,0,0}{tweeted}} {\colorbox[cmyk]{0,0,0,0}{:}} {\colorbox[cmyk]{0,0,0,0}{`}} {\colorbox[cmyk]{0,0,0,0}{looking}} {\colorbox[cmyk]{0,0,0,0}{forward}} {\colorbox[cmyk]{0,0,0,0}{to}} {\colorbox[cmyk]{0,0,0,0}{meeting}} {\colorbox[cmyk]{0,0,0,0}{@pele}} {\colorbox[cmyk]{0,0,0,0}{tonight}} {\colorbox[cmyk]{0,0,0,0}{.}} {\colorbox[cmyk]{0,0,0,0}{i}} {\colorbox[cmyk]{0,0,0,0}{ll}} {\colorbox[cmyk]{0,0,0,0}{be}} {\colorbox[cmyk]{0,0,0,0}{asking}} {\colorbox[cmyk]{0,0,0,0}{the}} {\colorbox[cmyk]{0,0,0,0}{best}} {\colorbox[cmyk]{0,0,0,0}{questions}} {\colorbox[cmyk]{0,0,0,0}{you}} {\colorbox[cmyk]{0,0,0,0}{sent}} {\colorbox[cmyk]{0,0,0,0}{.}} {\colorbox[cmyk]{0,0,0,0}{\#jackmeetspele}} {\colorbox[cmyk]{0,0,0,0}{.}} {\colorbox[cmyk]{0,0,0,0}{}} {\colorbox[cmyk]{0,0,0,0}{the}} {\colorbox[cmyk]{0,0,0,0}{23-year-old}} {\colorbox[cmyk]{0,0,0,0}{then}} {\colorbox[cmyk]{0,0,0,0}{followed}} {\colorbox[cmyk]{0,0,0,0}{this}} {\colorbox[cmyk]{0,0,0,0}{up}} {\colorbox[cmyk]{0,0,0,0}{with}} {\colorbox[cmyk]{0,0,0,0}{several}} {\colorbox[cmyk]{0,0,0,0}{tweets}} {\colorbox[cmyk]{0,0,0,0}{about}} {\colorbox[cmyk]{0,0,0,0}{the}} {\colorbox[cmyk]{0,0,0,0}{event}} {\colorbox[cmyk]{0,0,0,0}{,}} {\colorbox[cmyk]{0,0,0,0}{many}} {\colorbox[cmyk]{0,0,0,0}{of}} {\colorbox[cmyk]{0,0,0,0}{which}} {\colorbox[cmyk]{0,0,0,0}{included}} {\colorbox[cmyk]{0,0,0,0}{photos}} {\colorbox[cmyk]{0,0,0,0}{of}} {\colorbox[cmyk]{0,0,0,0}{pele}} {\colorbox[cmyk]{0,0,0,0}{.}} {\colorbox[cmyk]{0,0,0,0}{meanwhile}} {\colorbox[cmyk]{0,0,0,0}{,}} {\colorbox[cmyk]{0,0,0,0}{pele}} {\colorbox[cmyk]{0,0,0,0}{has}} {\colorbox[cmyk]{0,0,0,0}{acknowledged}} {\colorbox[cmyk]{0,0,0,0}{that}} {\colorbox[cmyk]{0,0,0,0}{last}} {\colorbox[cmyk]{0,0,0,0}{year}} {\colorbox[cmyk]{0,0,0,0}{s}} {\colorbox[cmyk]{0,0,0,0}{world}} {\colorbox[cmyk]{0,0,0,0}{cup}} {\colorbox[cmyk]{0,0,0,0}{was}} {\colorbox[cmyk]{0,0,0,0}{a}} {\colorbox[cmyk]{0,0,0,0}{`}} {\colorbox[cmyk]{0,0,0,0}{disaster}} {\colorbox[cmyk]{0,0,0,0}{}} {\colorbox[cmyk]{0,0,0,0}{for}} {\colorbox[cmyk]{1.0,0,1.0,0.07}{\doubleunderline{brazil}}} {\colorbox[cmyk]{1.0,0,1.0,0.08}{\doubleunderline{but}}} {\colorbox[cmyk]{1.0,0,1.0,0.19}{\doubleunderline{is}}} {\colorbox[cmyk]{1.0,0,1.0,0}{\doubleunderline{not}}} {\colorbox[cmyk]{1.0,0,1.0,0.04}{\doubleunderline{surprised}}} {\colorbox[cmyk]{1.0,0,1.0,0.05}{\doubleunderline{how}}} {\colorbox[cmyk]{1.0,0,1.0,0.11}{\doubleunderline{quickly}}} {\colorbox[cmyk]{1.0,0,1.0,0.05}{\doubleunderline{the}}} {\colorbox[cmyk]{1.0,0,1.0,0.1}{\doubleunderline{likes}}} {\colorbox[cmyk]{1.0,0,1.0,0.02}{\doubleunderline{of}}} {\colorbox[cmyk]{1.0,0,1.0,0.14}{\doubleunderline{oscar}}} {\colorbox[cmyk]{1.0,0,1.0,0.01}{\doubleunderline{and}}} {\colorbox[cmyk]{1.0,0,1.0,0}{\doubleunderline{ramires}}} {\colorbox[cmyk]{1.0,0,1.0,0.05}{\doubleunderline{have}}} {\colorbox[cmyk]{1.0,0,1.0,0.01}{\doubleunderline{bounced}}} {\colorbox[cmyk]{1.0,0,1.0,0.01}{\doubleunderline{back}}} {\colorbox[cmyk]{1.0,0,1.0,0.01}{\doubleunderline{in}}} {\colorbox[cmyk]{1.0,0,1.0,0.25}{\doubleunderline{the}}} {\colorbox[cmyk]{0,0,0,0}{barclays}} {\colorbox[cmyk]{1.0,0,1.0,0.4}{\doubleunderline{premier}}} {\colorbox[cmyk]{1.0,0,1.0,0.16}{\doubleunderline{league}}} {\colorbox[cmyk]{1.0,0,1.0,0}{\doubleunderline{this}}} {\colorbox[cmyk]{1.0,0,1.0,0.03}{\doubleunderline{season}}} {\colorbox[cmyk]{1.0,0,1.0,0.13}{\doubleunderline{.}}} {\colorbox[cmyk]{0,0,0,0}{brazil}} {\colorbox[cmyk]{0,0,0,0}{were}} {\colorbox[cmyk]{0,0,0,0}{humiliated}} {\colorbox[cmyk]{0,0,0,0}{by}} {\colorbox[cmyk]{0,0,0,0}{germany}} {\colorbox[cmyk]{0,0,0,0}{in}} {\colorbox[cmyk]{0,0,0,0}{a}} {\colorbox[cmyk]{0,0,0,0}{7-1}} {\colorbox[cmyk]{0,0,0,0}{semi-final}} {\colorbox[cmyk]{0,0,0,0}{defeat}} {\colorbox[cmyk]{0,0,0,0}{and}} {\colorbox[cmyk]{0,0,0,0}{the}} {\colorbox[cmyk]{0,0,0,0}{hosts}} {\colorbox[cmyk]{0,0,0,0}{were}} {\colorbox[cmyk]{0,0,0,0}{then}} {\colorbox[cmyk]{0,0,0,0}{thrashed}} {\colorbox[cmyk]{0,0,0,0}{3-0}} {\colorbox[cmyk]{0,0,0,0}{by}} {\colorbox[cmyk]{0,0,0,0}{holland}} {\colorbox[cmyk]{0,0,0,0}{in}} {\colorbox[cmyk]{0,0,0,0}{the}} {\colorbox[cmyk]{0,0,0,0}{third-place}} {\colorbox[cmyk]{0,0,0,0}{play-off}} {\colorbox[cmyk]{0,0,0,0}{.}} {\colorbox[cmyk]{0,0,0,0}{pele}} {\colorbox[cmyk]{0,0,0,0}{scored}} {\colorbox[cmyk]{1.0,0,1.0,0.54}{\doubleunderline{77}}} {\colorbox[cmyk]{1.0,0,1.0,0.42}{\doubleunderline{goals}}} {\colorbox[cmyk]{1.0,0,1.0,0.28}{\doubleunderline{in}}} {\colorbox[cmyk]{1.0,0,1.0,0.3}{\doubleunderline{92}}} {\colorbox[cmyk]{1.0,0,1.0,0.3}{\doubleunderline{games}}} {\colorbox[cmyk]{1.0,0,1.0,0.19}{\doubleunderline{for}}} {\colorbox[cmyk]{1.0,0,1.0,0.11}{\doubleunderline{brazil}}} {\colorbox[cmyk]{1.0,0,1.0,0.14}{\doubleunderline{and}}} {\colorbox[cmyk]{1.0,0,1.0,0.35}{\doubleunderline{won}}} {\colorbox[cmyk]{0,0,0,0}{the}} {\colorbox[cmyk]{1.0,0,1.0,0.55}{\doubleunderline{world}}} {\colorbox[cmyk]{0,0,0,0}{cup}} {\colorbox[cmyk]{0,0,0,0}{three}} {\colorbox[cmyk]{0,0,0,0}{times}} {\colorbox[cmyk]{0,0,0,0}{but}} {\colorbox[cmyk]{0,0,0,0}{the}} {\colorbox[cmyk]{0,0,0,0}{former}} {\colorbox[cmyk]{0,0,0,0}{santos}} {\colorbox[cmyk]{0,0,0,0}{striker}} {\colorbox[cmyk]{0,0,0,0}{still}} {\colorbox[cmyk]{0,0,0,0}{finds}} {\colorbox[cmyk]{0,0,0,0}{last}} {\colorbox[cmyk]{0,0,0,0}{year}} {\colorbox[cmyk]{0,0,0,0}{s}} {\colorbox[cmyk]{0,0,0,0}{capitulation}} {\colorbox[cmyk]{0,0,0,0}{difficult}} {\colorbox[cmyk]{0,0,0,0}{to}} {\colorbox[cmyk]{0,0,0,0}{understand}} {\colorbox[cmyk]{0,0,0,0}{.}}
\\
\hline
\textbf{Target Summary:}\\
\textbf{jack wilshere} was joined by former \textbf{england} manager \textbf{glenn hoddle}. the \textbf{arsenal} midfielder interviewed \textbf{pele} at launch of \textbf{10ten talent}. \textbf{pele} scored 77 goals in 92 games for \textbf{brazil} and won three world cups. the \textbf{brazil} legend says the 2014 world cup performance was not expected. the hosts were humiliated 7-1 by \textbf{germany} in the semi-finals last summer. \textbf{pele} is, however, not surprised by reaction of \textbf{oscar} and \textbf{ramires} this year.\\
\hline
\textbf{Baseline Model Prediction:}\\
jack wilshere was given the opportunity to interview the three-time world cup winner. both wilshere and pele are clients and the england international. pele has acknowledged that last year's world cup was a `disaster'
\\
\hline
\textbf{Our Model Prediction:}
\\
jack wilshere was given the `honour to interview the legendary pele' and asked twitter questions from fans. pele has acknowledged that last year's world cup was a `disaster' for brazil but is not surprised how quickly the likes of oscar and ramires have bounced back in the premier league this season. the brazil legend scored 77 goals in 92 games for brazil and won the world cup three times.
\\
\hline
\end{tabular}
\end{center}
\caption{Example 4134 from the \textit{CNN/Daily Mail} test set. Colors and underlines in the source reflect differences between baseline and our model attention weights: Red and a single underline reflects words attended by baseline model and not our model, Green and double underline reflects the opposite. Entities in bold in the target summary are answers to the example questions.}
\label{figure:example_with_attention}
\end{figure*}

We report our results in Table~\ref{table:apes_table}. For each system, we present its APES score alongside its F1 scores for ROUGE-1, ROUGE-2 and ROUGE-L, computed using  pyrouge \footnote{\url{https://pypi.org/project/pyrouge/}}.

We first report APES results on full source documents and gold summaries, in order to assess the capabilities of the QA system used for APES.  A simple answer extractor could answer 100\% of the questions given the gold-summaries. But the QA system is trained over the source documents and learns to generalize and not ``just" extract the answer. Answering questions from the full documents is indeed more difficult than from the gold-summaries because the QA system must locate the answer among multiple distractors. While gold-summaries present a very high APES score, the score reported for the source documents (61.1\%) is a realistic upper bound for APES.

We then present shuffled gold-summaries, where we randomly shuffled the location of each unigram in the gold summary. This score shows that even when all salient entities are in the shuffled text, APES is sensitive to the loss of coherence, readability and meaning. This confirms that APES does not only match the presence of entities. In contrast, ROUGE-1 fails to punish such incoherent sequences. Finally, we report ROUGE and APES for the strong Lead~3 sentences of the source document - a baseline known to beat most existing abstractive methods.

We then present APES and ROUGE scores for abstractive models, \citet{see2017get}'s model, our baseline model and our APES-optimized model. Our model achieves significantly higher APES scores (46.1 vs. 39.8) and improves all ROUGE metrics (by about 1 F-point over the baselines). The scores on the validation set are 46.6, 41.2, 18.4, 38.1 for APES, R1, R2, RL respectively.

While our objective is maximizing APES score, our model also increases its corresponding ROUGE scores. Unlike \citet{paulus2017deep} where the authors suggested a Reinforcement Learning based model to optimize ROUGE specifically, we optimize for APES and gain better ROUGE score.

We finally report the results obtained by our model when gold salient entities positions are given as oracle inputs instead of the predicted $a^e$ scores. The corresponding score (46.3 vs. 46.1) is only slightly above the score obtained by our model.  This indicates that the component of our model predicting entity saliency is good enough to drive summarization.

We carried out an informal error analysis to examine why some summaries perform worse than others with our architecture. We compared summaries that produce perfect APES score (1,630 out of 11,490 total) to the summaries with zero APES score (1,691). We measure the density of salient named entities in the source document: \#(salient entity mentions)/\#(distinct salient entities).
This density in the case of perfect APES summaries is much higher than that for low APES summaries (4.9 vs. 3.6). This observation suggests that we fail to produce higher APES scores when the salient entities aren't marked through sheer repetition.
\section{Conclusion}

We introduced APES, a new automatic summarization evaluation metric for news articles datasets based on the ability of a summary to answer questions regarding salient information from the text. 
This approach is useful in domains with source documents of about 1k words that focus on named entities - such as news articles, where named entities are effectively aligned with Pyramid SCUs. In other non-news domains, and longer documents, other methods for generating questions should be designed.  We compare APES to manual evaluation metrics on the TAC 2011 AESOP task and confirm its value as a complement to ROUGE. 

We introduce a new abstractive model that optimizes APES scores on the \textit{CNN/Daily Mail} dataset by attending salient entities from the input document, which also provides competitive ROUGE scores.
\section*{Acknowledgements}

This research was supported by the Lynn and William Frankel Centre for Computer Science at Ben-Gurion University.

\bibliography{naaclhlt2019}

\begin{thebibliography}{35}
\expandafter\ifx\csname natexlab\endcsname\relax\def\natexlab#1{#1}\fi

\bibitem[{Bahdanau et~al.(2014)Bahdanau, Cho, and Bengio}]{bahdanau2014neural}
Dzmitry Bahdanau, Kyunghyun Cho, and Yoshua Bengio. 2014.
\newblock Neural machine translation by jointly learning to align and
  translate.
\newblock \emph{arXiv preprint arXiv:1409.0473}.

\bibitem[{Celikyilmaz et~al.(2018)Celikyilmaz, Bosselut, He, and
  Choi}]{celikyilmaz2018deep}
Asli Celikyilmaz, Antoine Bosselut, Xiaodong He, and Yejin Choi. 2018.
\newblock Deep communicating agents for abstractive summarization.
\newblock \emph{arXiv preprint arXiv:1803.10357}.

\bibitem[{Chen et~al.(2016)Chen, Bolton, and Manning}]{chen2016thorough}
Danqi Chen, Jason Bolton, and Christopher~D Manning. 2016.
\newblock A thorough examination of the cnn/daily mail reading comprehension
  task.
\newblock \emph{arXiv preprint arXiv:1606.02858}.

\bibitem[{Cheng and Lapata(2016)}]{cheng2016neural}
Jianpeng Cheng and Mirella Lapata. 2016.
\newblock Neural summarization by extracting sentences and words.
\newblock \emph{arXiv preprint arXiv:1603.07252}.

\bibitem[{Dang(2005)}]{respon}
Hoa~Trang Dang. 2005.
\newblock Overview of duc 2005.
\newblock In \emph{Proceedings of the document understanding conference},
  volume 2005, pages 1--12.

\bibitem[{Duchi et~al.(2011)Duchi, Hazan, and Singer}]{duchi2011adaptive}
John Duchi, Elad Hazan, and Yoram Singer. 2011.
\newblock Adaptive subgradient methods for online learning and stochastic
  optimization.
\newblock \emph{Journal of Machine Learning Research}, 12(Jul):2121--2159.

\bibitem[{Gehrmann et~al.(2018)Gehrmann, Deng, and Rush}]{gehrmann2018bottom}
Sebastian Gehrmann, Yuntian Deng, and Alexander~M Rush. 2018.
\newblock Bottom-up abstractive summarization.
\newblock \emph{arXiv preprint arXiv:1808.10792}.

\bibitem[{Graff et~al.(2003)Graff, Kong, Chen, and Maeda}]{graff2003english}
David Graff, Junbo Kong, Ke~Chen, and Kazuaki Maeda. 2003.
\newblock English gigaword.
\newblock \emph{Linguistic Data Consortium, Philadelphia}, 4:1.

\bibitem[{Hermann et~al.(2015)Hermann, Kocisky, Grefenstette, Espeholt, Kay,
  Suleyman, and Blunsom}]{hermann2015teaching}
Karl~Moritz Hermann, Tomas Kocisky, Edward Grefenstette, Lasse Espeholt, Will
  Kay, Mustafa Suleyman, and Phil Blunsom. 2015.
\newblock Teaching machines to read and comprehend.
\newblock In \emph{Advances in Neural Information Processing Systems}, pages
  1693--1701.

\bibitem[{Hirao et~al.(2018)Hirao, Kamigaito, and
  Nagata}]{Hirao2018AutomaticPE}
Tsutomu Hirao, Hidetaka Kamigaito, and Masaaki Nagata. 2018.
\newblock Automatic pyramid evaluation exploiting edu-based extractive
  reference summaries.
\newblock In \emph{EMNLP}.

\bibitem[{Hobson et~al.(2007)Hobson, Dorr, Monz, and Schwartz}]{hobson2007task}
Stacy~President Hobson, Bonnie~J Dorr, Christof Monz, and Richard Schwartz.
  2007.
\newblock Task-based evaluation of text summarization using relevance
  prediction.
\newblock \emph{Information Processing \& Management}, 43(6):1482--1499.

\bibitem[{Honnibal and Johnson(2015)}]{spacy}
Matthew Honnibal and Mark Johnson. 2015.
\newblock \href {https://aclweb.org/anthology/D/D15/D15-1162} {An improved
  non-monotonic transition system for dependency parsing}.
\newblock In \emph{Proceedings of the 2015 Conference on Empirical Methods in
  Natural Language Processing}, pages 1373--1378, Lisbon, Portugal. Association
  for Computational Linguistics.

\bibitem[{Hovy et~al.(2006)Hovy, Lin, Zhou, and Fukumoto}]{hovy2006automated}
Eduard Hovy, Chin-Yew Lin, Liang Zhou, and Junichi Fukumoto. 2006.
\newblock Automated summarization evaluation with basic elements.
\newblock In \emph{Proceedings of the Fifth Conference on Language Resources
  and Evaluation (LREC 2006)}, pages 604--611. Citeseer.

\bibitem[{Jing et~al.(1998)Jing, Barzilay, McKeown, and
  Elhadad}]{jing1998summarization}
Hongyan Jing, Regina Barzilay, Kathleen McKeown, and Michael Elhadad. 1998.
\newblock Summarization evaluation methods: Experiments and analysis.
\newblock In \emph{AAAI symposium on intelligent summarization}, pages 51--59.

\bibitem[{Klein et~al.(2017)Klein, Kim, Deng, Senellart, and Rush}]{opennmt}
Guillaume Klein, Yoon Kim, Yuntian Deng, Jean Senellart, and Alexander~M. Rush.
  2017.
\newblock \href {https://doi.org/10.18653/v1/P17-4012} {Open{NMT}: Open-source
  toolkit for neural machine translation}.
\newblock In \emph{Proc. ACL}.

\bibitem[{Lin(2004)}]{rouge}
Chin-Yew Lin. 2004.
\newblock Rouge: A package for automatic evaluation of summaries.
\newblock In \emph{Text summarization branches out: Proceedings of the ACL-04
  workshop}, volume~8. Barcelona, Spain.

\bibitem[{Louis and Nenkova(2013)}]{louis2013automatically}
Annie Louis and Ani Nenkova. 2013.
\newblock Automatically assessing machine summary content without a gold
  standard.
\newblock \emph{Computational Linguistics}, 39(2):267--300.

\bibitem[{Nallapati et~al.(2017)Nallapati, Zhai, and
  Zhou}]{nallapati2017summarunner}
Ramesh Nallapati, Feifei Zhai, and Bowen Zhou. 2017.
\newblock Summarunner: A recurrent neural network based sequence model for
  extractive summarization of documents.
\newblock \emph{hiP (yi= 1| hi, si, d)}, 1:1.

\bibitem[{Nallapati et~al.(2016)Nallapati, Zhou, Gulcehre, Xiang
  et~al.}]{nallapati2016abstractive}
Ramesh Nallapati, Bowen Zhou, Caglar Gulcehre, Bing Xiang, et~al. 2016.
\newblock Abstractive text summarization using sequence-to-sequence rnns and
  beyond.
\newblock \emph{arXiv preprint arXiv:1602.06023}.

\bibitem[{Narayan et~al.(2018)Narayan, Cohen, and Lapata}]{narayan2018ranking}
Shashi Narayan, Shay~B Cohen, and Mirella Lapata. 2018.
\newblock Ranking sentences for extractive summarization with reinforcement
  learning.
\newblock \emph{arXiv preprint arXiv:1802.08636}.

\bibitem[{Nenkova et~al.(2007)Nenkova, Passonneau, and McKeown}]{pyramid}
Ani Nenkova, Rebecca Passonneau, and Kathleen McKeown. 2007.
\newblock The pyramid method: Incorporating human content selection variation
  in summarization evaluation.
\newblock \emph{ACM Transactions on Speech and Language Processing (TSLP)},
  4(2):4.

\bibitem[{Owczarzak(2009)}]{Owczarzak2009DEPEVALsummDE}
Karolina Owczarzak. 2009.
\newblock Depeval(summ): Dependency-based evaluation for automatic summaries.
\newblock In \emph{ACL/IJCNLP}.

\bibitem[{Owczarzak and Dang(2011)}]{tac11}
Karolina Owczarzak and Hoa~Trang Dang. 2011.
\newblock Overview of the tac 2011 summarization track: Guided task and aesop
  task.
\newblock In \emph{Proceedings of the Text Analysis Conference (TAC 2011),
  Gaithersburg, Maryland, USA, November}.

\bibitem[{Paszke et~al.(2017)Paszke, Gross, Chintala, Chanan, Yang, DeVito,
  Lin, Desmaison, Antiga, and Lerer}]{paszke2017automatic}
Adam Paszke, Sam Gross, Soumith Chintala, Gregory Chanan, Edward Yang, Zachary
  DeVito, Zeming Lin, Alban Desmaison, Luca Antiga, and Adam Lerer. 2017.
\newblock Automatic differentiation in pytorch.
\newblock In \emph{NIPS-W}.

\bibitem[{Paulus et~al.(2017)Paulus, Xiong, and Socher}]{paulus2017deep}
Romain Paulus, Caiming Xiong, and Richard Socher. 2017.
\newblock A deep reinforced model for abstractive summarization.
\newblock \emph{arXiv preprint arXiv:1705.04304}.

\bibitem[{Rush et~al.(2015)Rush, Chopra, and Weston}]{rush2015neural}
Alexander~M Rush, Sumit Chopra, and Jason Weston. 2015.
\newblock A neural attention model for abstractive sentence summarization.
\newblock \emph{arXiv preprint arXiv:1509.00685}.

\bibitem[{Sandhaus(2008)}]{sandhaus2008new}
Evan Sandhaus. 2008.
\newblock The new york times annotated corpus.
\newblock \emph{Linguistic Data Consortium, Philadelphia}, 6(12):e26752.

\bibitem[{See et~al.(2017)See, Liu, and Manning}]{see2017get}
Abigail See, Peter~J Liu, and Christopher~D Manning. 2017.
\newblock Get to the point: Summarization with pointer-generator networks.
\newblock \emph{arXiv preprint arXiv:1704.04368}.

\bibitem[{Steinberger and Je{\v{z}}ek(2012)}]{steinberger2012evaluation}
Josef Steinberger and Karel Je{\v{z}}ek. 2012.
\newblock Evaluation measures for text summarization.
\newblock \emph{Computing and Informatics}, 28(2):251--275.

\bibitem[{Sutskever et~al.(2014)Sutskever, Vinyals, and
  Le}]{sutskever2014sequence}
Ilya Sutskever, Oriol Vinyals, and Quoc~V Le. 2014.
\newblock Sequence to sequence learning with neural networks.
\newblock In \emph{Advances in neural information processing systems}, pages
  3104--3112.

\bibitem[{Tay et~al.(2017)Tay, Phan, Tuan, and Hui}]{tay2017skipflow}
Yi~Tay, Minh~C Phan, Luu~Anh Tuan, and Siu~Cheung Hui. 2017.
\newblock Skipflow: Incorporating neural coherence features for end-to-end
  automatic text scoring.
\newblock \emph{arXiv preprint arXiv:1711.04981}.

\bibitem[{Vadlapudi and Katragadda(2010)}]{vadlapudi2010automated}
Ravikiran Vadlapudi and Rahul Katragadda. 2010.
\newblock On automated evaluation of readability of summaries: Capturing
  grammaticality, focus, structure and coherence.
\newblock In \emph{Proceedings of the NAACL HLT 2010 student research
  workshop}, pages 7--12. Association for Computational Linguistics.

\bibitem[{Vinyals et~al.(2015)Vinyals, Fortunato, and
  Jaitly}]{vinyals2015pointer}
Oriol Vinyals, Meire Fortunato, and Navdeep Jaitly. 2015.
\newblock Pointer networks.
\newblock In \emph{Advances in Neural Information Processing Systems}, pages
  2692--2700.

\bibitem[{Wu et~al.(2016)Wu, Schuster, Chen, Le, Norouzi, Macherey, Krikun,
  Cao, Gao, Macherey et~al.}]{wu2016google}
Yonghui Wu, Mike Schuster, Zhifeng Chen, Quoc~V Le, Mohammad Norouzi, Wolfgang
  Macherey, Maxim Krikun, Yuan Cao, Qin Gao, Klaus Macherey, et~al. 2016.
\newblock Google's neural machine translation system: Bridging the gap between
  human and machine translation.
\newblock \emph{arXiv preprint arXiv:1609.08144}.

\bibitem[{Yang et~al.(2016)Yang, Passonneau, and de~Melo}]{Yang2016PEAKPE}
Qian Yang, Rebecca~J. Passonneau, and Gerard de~Melo. 2016.
\newblock Peak: Pyramid evaluation via automated knowledge extraction.
\newblock In \emph{AAAI}.

\end{thebibliography}
\bibliographystyle{acl_natbib}

\appendix
\section{Experiment Settings}

For our experiments, we used a bidirectional LSTM encoder with 256-dimensional hidden states for each direction, an LSTM decoder with 512-dimensional hidden states and 128-dimensional embeddings for a 50k shared-vocabulary words. We do not use pretrained word embeddings. 

We use the Adagrad \cite{duchi2011adaptive} optimizer with a starting learning rate of $0.15$ and gradient clipping with a maximum gradient norm of 2. At train-time source and target documents are truncated to 400 and 100 tokens respectively. After training our baseline model for 20 epochs, we fine-tune the network with Eq. \eqref{eqn:hybrid_loss} loss for an additional 5 epochs starting again with 0.15 as initial learning rate. Results reported in this paper correspond to $\lambda=0.01$.

At test-time, we do not truncate the source documents enabling the network to attend overall input text. We use Eq.~\eqref{eqn:my_beam_pen} as the beam search score function, penalizing using $cp(X;Y)$ every single decoding step and $lp(Y)$ and $ep(X;Y)$ only when all hypotheses are done. We choose $\alpha, \beta, \gamma$ values of $0.9, 0.5, 0.5$ respectively for our model. We also used \citet{paulus2017deep} suggestion of repetition avoidance by blocking trigrams appearing more than once at inference time.

Running APES evaluation on a generated test set (of size 11,490 summaries) takes about 40 minutes using a single process.

\end{document}